# SAVERS: SAR ATR with Verification Support Based on Convolutional Neural Network

Hidetoshi FURUKAWA

1778-2, Furuichiba, Saiwai-ku, Kawasaki-shi, Kanagawa, 212–0052 Japan
E-mail: hidetoshi.furukawa@ai4sig.org

**Abstract**　We propose a new convolutional neural network (CNN) which performs coarse and fine segmentation for end-to-end synthetic aperture radar (SAR) automatic target recognition (ATR) system. In recent years, many CNNs for SAR ATR using deep learning have been proposed, but most of them classify target classes from fixed size target chips extracted from SAR imagery. On the other hand, we proposed the CNN which outputs the score of the multiple target classes and a background class for each pixel from the SAR imagery of arbitrary size and multiple targets as fine segmentation. However, it was necessary for humans to judge the CNN segmentation result. In this report, we propose a CNN called SAR ATR with verification support (SAVERS), which performs region-wise (i.e. coarse) segmentation and pixel-wise segmentation. SAVERS discriminates between target and non-target, and classifies multiple target classes and non-target class by coarse segmentation. This report describes the evaluation results of SAVERS using the Moving and Stationary Target Acquisition and Recognition (MSTAR) dataset.

**Key words**　Automatic target recognition (ATR), Detection, Discrimination, Classification, Convolutional neural network (CNN), Synthetic aperture radar (SAR)

## 1. Introduction

In recent years, methods using convolution neural network (CNN) [1]–[4] have been successful in the classification of image recognition. Similarly, CNNs for synthetic aperture radar (SAR) automatic target recognition (ATR) have been proposed. On the Moving and Stationary Target Acquisition and Recognition (MSTAR) public dataset [5], the target classification accuracy of the CNNs [6]–[9] exceeds conventional methods (support vector machine, etc.). However, most of CNNs for SAR ATR classify target classes from a target chip extracted from SAR image but do not classify multiple targets or a target chip (or SAR image) of an arbitrary size. In addition, a CNN for target classification can output score or probability of each class as classification result, but it is difficult for a human to verify the classification result.

Figure 2a shows that the standard architecture of SAR ATR consists of three stages: detection, discrimination, and classification. Detection: the first stage of SAR ATR detects a region of interest (ROI) from a SAR image. Discrimination: the second stage of SAR ATR discriminates whether an ROI is a target or non-target region, and outputs the discriminated ROI as a target chip. Classification: the third stage of SAR ATR classifies target classes from a target chip.

In contrast, we proposed an architecture that performs detection, discrimination, and classification in a single stage

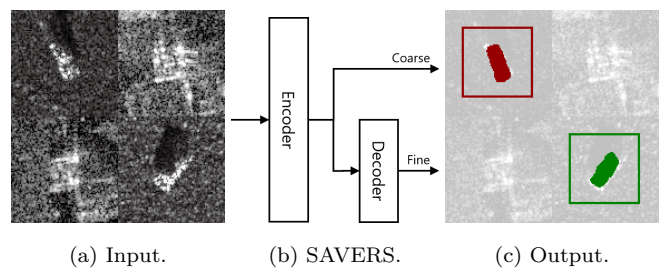

(a) Input.　(b) SAVERS.　(c) Output.

Fig. 1　Illustration of input and output of proposed CNN. The CNN named SAVERS performs automatic target recognition of multi-class / multi-target in variable size SAR image. In this case, the input is a single image with two targets of different classes and two clutters. SAVERS outputs the position, class, and shape of each detected target.

(Fig. 2b). Furthermore, we proposed a CNN which inputs a SAR image of variable sizes with multi-target and outputs a SAR ATR image.

In this report, we propose a new CNN focusing on object detection by coarse segmentation and discrimination between target and non-target using clutter chips.

## 2. Related Work

In segmentation of image recognition giving classification label for each pixel of an image, methods using CNN [10]–[12] show a good performance in recent years. For SAR image, segmentation of a target region and a shadow region is performed. The CNN [13] and other methods [14]–[17] have been proposed for target and shadow region segmentation of a SAR image. In contrast, we proposed the CNN named Versnet [18] performs target detection, target classification, and pose estimation by segmentation.

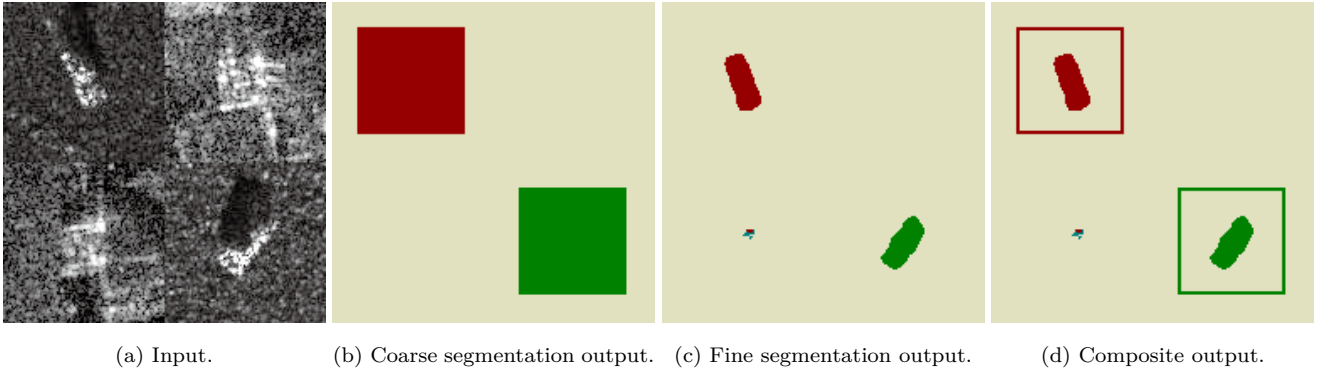

(a) Input.    (b) Coarse segmentation output.    (c) Fine segmentation output.    (d) Composite output.

Fig. 3   Outputs of coarse segmentation, fine segmentation, and composite.

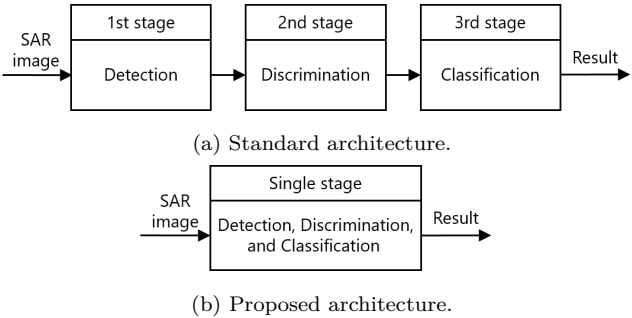

(a) Standard architecture.

(b) Proposed architecture.

Fig. 2   Architecture for an end-to-end SAR ATR system. The standard architecture is split into three stages. The Proposed architecture consists of a single stage.

Similarly, object detection that estimates the position, size, and target class of multi-target is also known. In object detection, the position and size are represented by bounding boxes. For object detection, CNNs [19]–[21] are proposed. Also, the CNN for object detection is applied to SAR ATR in [22].

On the other hand, we propose a CNN which simultaneously performs object detection and segmentation.

## 3. Proposed Method

A proposed CNN named SAR ATR with verification support (SAVERS) inputs an arbitrary size SAR image with multiple classes and multiple targets, and outputs the position, class, and shape of each detected target as a SAR ATR image.

Figure 1 shows the outline of SAVERS for end-to-end SAR ATR. SAVERS is a CNN composed of an encoder and a decoder. The encoder of SAVERS extracts features from an input SAR image. The decoder converts the features based on the conversion rule in the training data and outputs it as a SAR ATR image.

Here, we define the end-to-end SAR ATR as a task of supervised learning. Let $\{(X_n, D_n), n = 1, ..., N\}$ be the training dataset, where $X_n = \{x_i^{(n)}, i = 1, ..., |X_n|\}$ is SAR image as input data, $D_n = \{d_i^{(n)}, i = 1, ..., |D_n|, d_i^{(n)} \in \{0, ..., N_c - 1\}\}$ is label image for $X_n$, which is the supervised data of SAVERS output data $Y_n = f(X_n; \theta)$. The values of $|X_n|$ and $|D_n|$ represent the number of pixels (vertical $\times$ horizontal) of SAR and label image, respectively. When $d_i^{(n)}$ is 0, it represents a background class, and when $d_i^{(n)}$ is 1 or more, it indicates a corresponding target class. Let $L(\theta)$ be a loss function, the network parameters $\theta$ are adjusted using training data so that the output of loss function becomes small.

## 4. Experiments

### 4.1 Dataset

For training and testing of SAVERS, we used the 11 classes data shown in Table 1. The image chips have ten target classes and a background (i.e. non-target) class.

The data of ten target classes contains 2747 target chips with a depression angle of 17° for the training and 2420 target chips with a depression angle of 15° for the testing from the MSTAR [5] dataset. Five target chips of target class BTR60 for testing data were excluded. Table 2 shows a list of target chips excluded from testing data.

The data of a background class contains 274 and 242 clutter chips for the training and testing, respectively. We use clutter chips provided by Adaptive SAR ATR Problem Set (AdaptSAPS) [23] using the MSTAR dataset.

Of course, label images for segmentation do not exist in the MSTAR dataset. Therefore, we create label images for SAVERS. Figure 4d shows samples of label images. The label images have all 11 classes: 10 target classes and a background (i.e. non-target) class.

### 4.2 SAVERS: Proposed CNN

Figure 5 shows a detailed architecture of SAVERS for ex-periments. The encoder of the SAVERS consists of four convolution blocks and two convolution layers. The convolution block contains two convolution layers of kernel size $3 \times 3$ and a max pooling layer similarly to VGG [24]. The activation function of all convolutions except the final convolution uses rectified linear unit (ReLU) [25]. Dropout [26] is applied after a convolution of kernel size $4 \times 4$. Batch normalization [27] is not applied. The decoder of the SAVERS consists of a transposed convolution [28] that performs 16 times upsampling.

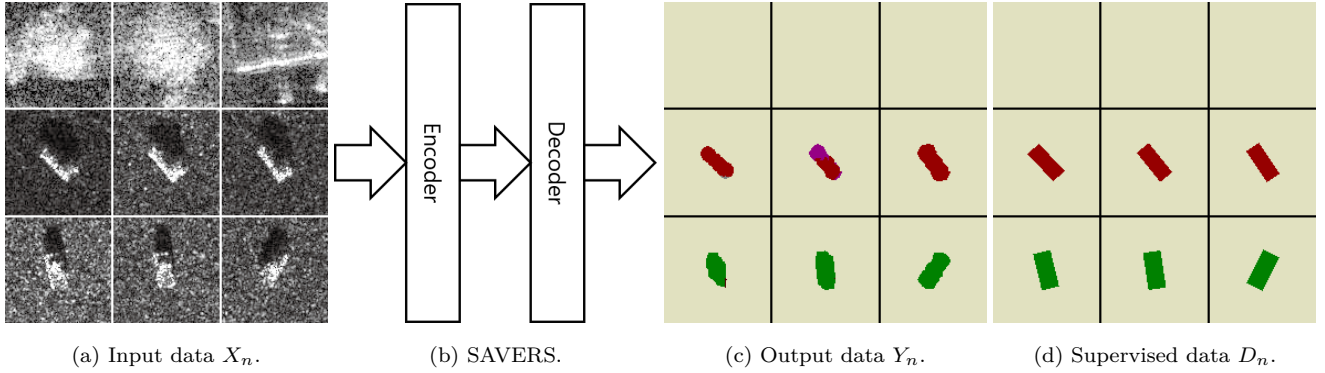

(a) Input data $X_n$.    (b) SAVERS.    (c) Output data $Y_n$.    (d) Supervised data $D_n$.

Fig. 4　Illustration of training for proposed CNN.

Table 1　Dataset. The training and testing data contain respectively 3021 and 2662 image chips.

| Class | Training data | Testing data |
|---|---|---|
| Background | 274 | 242 |
| 2S1 | 299 | 274 |
| BMP2 (9563) | 233 | 195 |
| BRDM2 | 298 | 274 |
| BTR60 | 256 | 190 |
| BTR70 | 233 | 196 |
| D7 | 299 | 274 |
| T62 | 299 | 273 |
| T72 (132) | 232 | 196 |
| ZIL131 | 299 | 274 |
| ZSU234 | 299 | 274 |
| Total | 3021 | 2662 |

Table 2　List of target chips excluded from testing data.

| Class | Filename | Aspect angle (°) |
|---|---|---|
| BTR60 | HB03353.003 | 305.48 |
| BTR60 | HB04933.003 | 303.48 |
| BTR60 | HB04999.003 | 299.48 |
| BTR60 | HB05000.003 | 304.48 |
| BTR60 | HB05631.003 | 302.48 |

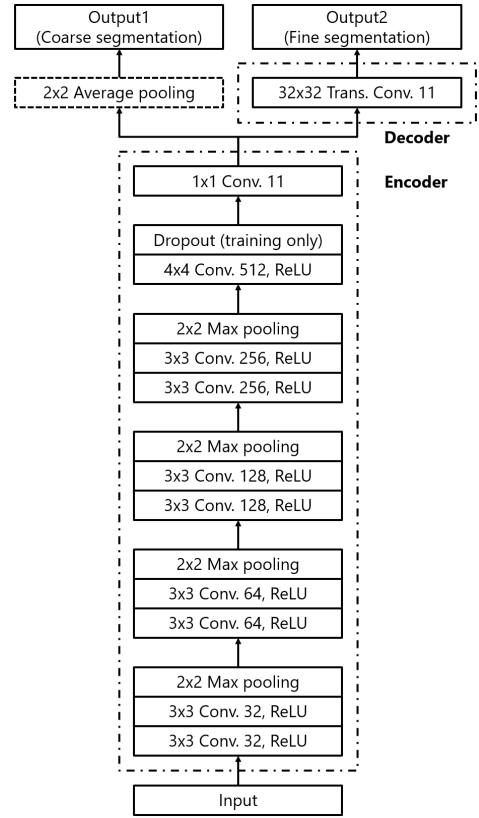

Fig. 5　Detail architecture of SAVERS for experiments. The SAVERS refers to the fully convolutional network called FCN-32s.

As the loss function, we use cross entropy expressed by

$$L(\theta) = -\sum_x p(x) \log q(x). \qquad (1)$$

For the optimization of the loss function, we use stochastic gradient descent (SGD) with momentum.

Since the SAVERS is a CNN without fully connected layers called fully convolutional network (FCN) [10], even if training is done with small size images, the SAVERS can process SAR images of arbitrary size.

### 4.3　Coarse Segmentation

First, we show results of coarse segmentation.

The decoder output performs pixel-wise segmentation (i.e. fine segmentation), whereas the encoder output is used for region-wise segmentation (i.e. coarse segmentation).

Figure 6a shows class accuracy using the encoder output before the average pooling layer. We simply use the maximum probability class (i.e. $\arg\max(p)$) as predicted class. The class accuracy before average pooling is high in the four areas near the center, but its value is 0.661 to 0.727. The class accuracy of the peripheral area is 0.091 (242/2662) because the clutter chips are correctly classified as background class.

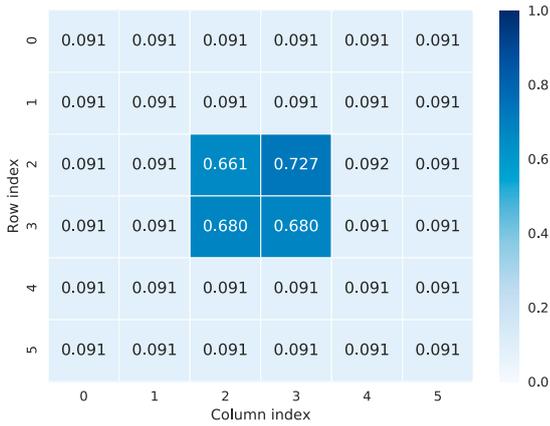

(a) Before average pooling.

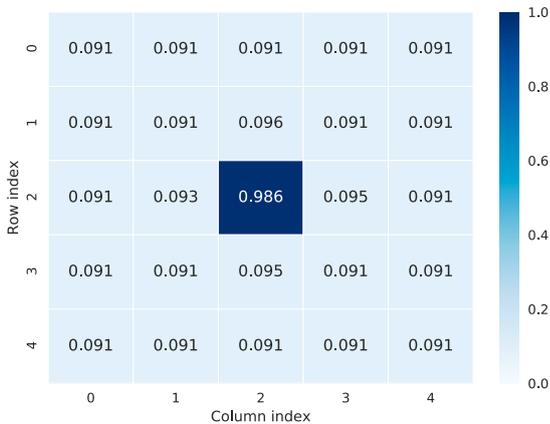

(b) After average pooling.

Fig. 6  Classification accuracy of coarse segmentation.

Figure 6b shows class accuracy using the encoder output after the average pooling layer. By performing an average pooling, the scores dispersed in the four regions are aggregated, and the class accuracy of the central region is increased to 0.986 (2624/2662).

### 4.4 Classification Performance of Coarse Segmentation

Next, we show results of classification performance of coarse segmentation output (i.e. the encoder output after the average pooling layer).

We use precision, recall, and $F_1$ as metrics of classification performance. Each metrics is given by

$$\text{Precision} = \frac{\text{TP}}{\text{TP} + \text{FP}}, \tag{2}$$

$$\text{Recall} = \frac{\text{TP}}{\text{TP} + \text{FN}}, \tag{3}$$

$$F_1 = 2 \cdot \frac{\text{precision} \cdot \text{recall}}{\text{precision} + \text{recall}}, \tag{4}$$

Table 3  Definitions of TP, FP, FN, and TN.

| Predicted \ True | Condition positive | Condition negative |
|---|---|---|
| Condition positive | True positive (TP) | False positive (FP) |
| Condition negative | False negative (FN) | True negative (TN) |

Table 4  Classification performance of testing.

| Class | Precision | Recall | $F_1$ |
|---|---|---|---|
| Background | 0.903 | 1.000 | 0.949 |
| 2S1 | 0.986 | 0.993 | 0.989 |
| BMP2 | 1.000 | 0.995 | 0.997 |
| BRDM2 | 1.000 | 0.978 | 0.989 |
| BTR60 | 1.000 | 0.974 | 0.987 |
| BTR70 | 1.000 | 0.990 | 0.995 |
| D7 | 1.000 | 0.960 | 0.980 |
| T62 | 0.989 | 0.985 | 0.987 |
| T72 | 0.995 | 1.000 | 0.997 |
| ZIL131 | 0.993 | 0.978 | 0.985 |
| ZSU234 | 0.993 | 0.996 | 0.995 |

where the definitions of TP, FP, FN, and TN are shown in Table 3.

Table 4 shows precision, recall, and $F_1$ of testing. Since all clutter chips are correctly classified as background (i.e. non-target) class, the recall is 1.000. However, because the part of the target chips is erroneously classified as background class, the precision becomes 0.903, and $F_1$ which is the harmonic average of precision and recall is 0.949.

Table 5 shows a confusion matrix for the image chips of testing. Each column in the confusion matrix represents the actual class, and each row represents the class predicted by the SAVERS.

Figure 7a shows a histogram of $(1 - p_0)$ for target and clutter chips, where $p_0$ is probability of background class obtained by softmax function.

Figure 7b shows a cumulative distribution of $(1 - p_0)$ for target chips. The empirical cumulative distribution function $P((1 - p_0) \leq 0.5)$ and $P((1 - p_0) \leq 0.8)$ are about 0.01 and 0.1, respectively.

### 4.5 Multi-Class and Multi-Target

Finally, we show the SAVERS output for multi-class and multi-target input. Figure 8 shows input (SAR image), output (SAR ATR image), and ground truth.

### 5. Conclusion

In this report, we proposed a CNN based on a new architecture consisting of a single stage for end-to-end SAR ATR system, not a standard architecture consisting of three stages. Unlike conventional CNN for target classification, the CNN named SAVERS inputs SAR imagery of arbitrary sizes

Table 5  Confusion matrix of testing data.

| Predicted \ True | Background | 2S1 | BMP2 | BRDM2 | BTR60 | BTR70 | D7 | T62 | T72 | ZIL131 | ZSU234 | Precision |
|---|---|---|---|---|---|---|---|---|---|---|---|---|
| Background | 242 | 1 | 0 | 3 | 3 | 1 | 11 | 0 | 0 | 6 | 1 | 0.903 |
| 2S1 | 0 | 272 | 1 | 1 | 0 | 1 | 0 | 1 | 0 | 0 | 0 | 0.986 |
| BMP2 | 0 | 0 | 194 | 0 | 0 | 0 | 0 | 0 | 0 | 0 | 0 | 1.000 |
| BRDM2 | 0 | 0 | 0 | 268 | 0 | 0 | 0 | 0 | 0 | 0 | 0 | 1.000 |
| BTR60 | 0 | 0 | 0 | 0 | 185 | 0 | 0 | 0 | 0 | 0 | 0 | 1.000 |
| BTR70 | 0 | 0 | 0 | 0 | 0 | 194 | 0 | 0 | 0 | 0 | 0 | 1.000 |
| D7 | 0 | 0 | 0 | 0 | 0 | 0 | 263 | 0 | 0 | 0 | 0 | 1.000 |
| T62 | 0 | 1 | 0 | 0 | 2 | 0 | 0 | 269 | 0 | 0 | 0 | 0.989 |
| T72 | 0 | 0 | 0 | 0 | 0 | 0 | 0 | 1 | 196 | 0 | 0 | 0.995 |
| ZIL131 | 0 | 0 | 0 | 2 | 0 | 0 | 0 | 0 | 0 | 268 | 0 | 0.993 |
| ZSU234 | 0 | 0 | 0 | 0 | 0 | 0 | 0 | 2 | 0 | 0 | 273 | 0.993 |
| Recall | 1.000 | 0.993 | 0.995 | 0.978 | 0.974 | 0.990 | 0.960 | 0.985 | 1.000 | 0.978 | 0.996 | |
| $F_1$ | 0.949 | 0.989 | 0.997 | 0.989 | 0.987 | 0.995 | 0.980 | 0.987 | 0.997 | 0.985 | 0.995 | |

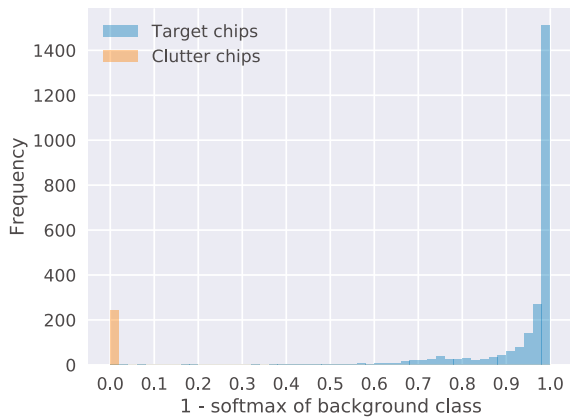

(a) Histogram.

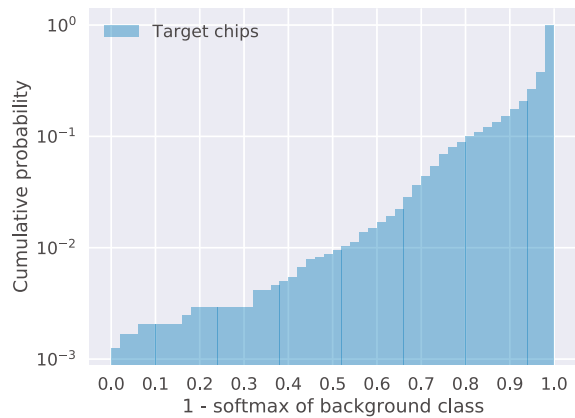

(b) Cumulative distribution.

Fig. 7  Histogram and cumulative distribution of (1 - softmax of background class).

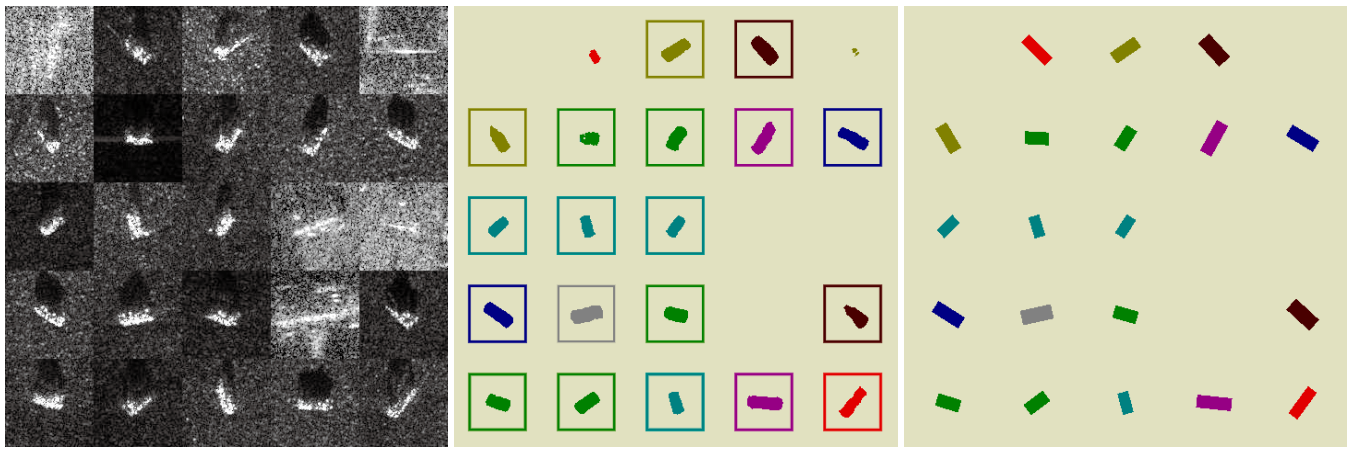

(a) Input (SAR image).   (b) Output (SAR ATR image).   (c) Ground truth.

Fig. 8  Multi-class and multi-target.

with multi-class and multi-target, discriminates between target and non-target (i.e. clutter), and output the position, class, and shape of each detected target as SAR ATR image.

We trained SAVERS to output scores of ten target classes and a background class (i.e. clutter) per pixel using the target chips of the MSTAR dataset and the clutter chips provided by AdaptSAPS, and we evaluated the performance of encoder output obtained by training. In the evaluation, the classification accuracy of the encoder output applied with average pooling was 98.6% (2624/2662).

Although this accuracy is inferior to the classification performance of state-of-the-art which classifies only the class of the target chip, it acquired the function of discrimination between target and non-target. A detailed analysis of the classification performance shows that clutter chips are correctly classified as background class, but part of the target chips are classified as background class, and extended study is future work.